\documentclass[letterpaper]{article} 
\usepackage[preprint]{aaai2027}
\usepackage[hyphens]{url}  
\usepackage{graphicx} 
\usepackage{natbib}  
\usepackage{caption} 
\usepackage{amsmath}
\usepackage{amssymb}
\usepackage{colortbl}

\title{TransPhy: Visual In-Context Learning for Physically Grounded Image Editing}

\author{
    Siyi Xie\textsuperscript{\rm 1,2},
    Xuanke Shi\textsuperscript{\rm 2},
    Jinsheng Quan\textsuperscript{\rm 2,3},
    Haoran Tang\textsuperscript{\rm 1},\\
    Zukai Chen\textsuperscript{\rm 2},
    Lei Yang\textsuperscript{\rm 2}\corresponding,
    Quan Wang\textsuperscript{\rm 2}\corresponding
}
\affiliations{
    \textsuperscript{\rm 1}Peking University
    \textsuperscript{\rm 2}SenseTime Research
    \textsuperscript{\rm 3}Zhejiang University
}

\begin{document}

\maketitle

\begin{abstract}
Visual demonstrations provide a natural interface for specifying image transformations that are difficult to describe exhaustively with text. However, existing visual in-context learning (VICL) methods primarily focus on appearance-level relation transfer and provide limited support for physically grounded transformations, whose outcomes depend on material properties, geometry, object interactions, and environmental conditions. Given a source--target exemplar pair and a query image, physically grounded VICL requires a model to infer the demonstrated transformation, adapt its effects to the query-specific scene context, and preserve rule-irrelevant content. We introduce PhysVICL-74, comprising 74 physically grounded transformation rules and 5,240 source--target image pairs that form nearly 75K training and evaluation contexts. Its benchmark split separately evaluates novel-instance transfer and unseen-rule generalization. We further propose TransPhy, a framework that decomposes physically grounded VICL into physical-rule induction and transition-aligned rendering. TransPhy first predicts the demonstrated rule and an explicit query-specific target-state description, and then synthesizes the target image through token-wise mixture-of-experts adaptation, with expert routing guided by localized transition cues. Experiments show that TransPhy improves physical-rule adherence, query consistency, and unseen-rule generalization over existing visual in-context editing methods.

\end{abstract}

\section{Introduction}
\begin{figure}[t]
    \centering
    \includegraphics[
        width=\columnwidth,
        trim=101bp 266.5bp 398.75bp 129.25bp,
        clip
    ]{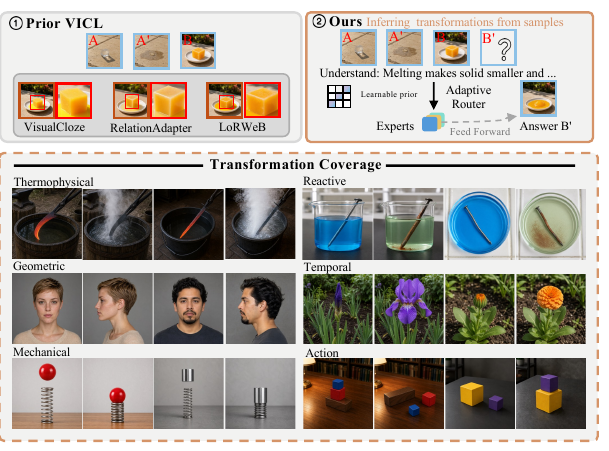}
    \caption{
        Physically grounded VICL. Top: limitations of prior methods (1) and our TransPhy framework (2). Bottom: representative PhysVICL-74 transformations.
    }
    \label{fig:task}
    \vspace{-10pt}
\end{figure}

Text prompts provide an intuitive interface for image editing \cite{brooks2023instructpix2pixlearningfollowimage, feng2025itemworthpromptversatile, samadi2024funeditorachievingcompleximage, feng2024dit4editdiffusiontransformerimage, mao2025tweezeeditconsistentefficientimage, ma2025x2editrevisitingarbitraryinstructionimage}. However, language often underspecifies transformations involving multiple coupled effects and complex spatial dependencies. Melting, for example, may require coordinated changes in material state, object shape, solid volume, and interaction with the supporting surface. Exhaustively describing these consequences in text is cumbersome and ambiguous. Visual in-context learning (VICL) provides a natural alternative: given a source--target exemplar pair, a model transfers the demonstrated transformation to a new query image. By showing both changed and preserved content, visual exemplars can specify transformations that are difficult to express precisely through language alone.

Existing VICL methods effectively transfer perceptual and semantic relations across stylization, visual effects, attribute editing, representation conversion, and image restoration \cite{li2025visualcloze,blackforestlabs2025fluxkontext,song2025insertanythingimageinsertion,zhang2025context,chen2025edittransfer,gong2025relationadapter,manor2026lorweb,rajagopalan2024awracleallweatherimagerestoration}. These tasks are largely evaluated by reproducing the demonstrated relation. Physically grounded transformations additionally require query-dependent adaptation: the same rule may yield different outcomes according to material, geometry, interactions, and environment. For example, melting ice and wax produces different deformations, material states, and secondary effects despite sharing a high-level rule. Successful transfer must therefore infer the transformation and adapt it to the query rather than copy the exemplar difference. We call a transformation \textit{physically grounded} when its observable outcome depends on these query-specific physical factors.

Given a source--target exemplar pair $(A,A')$ and a query image $B$, physically grounded VICL synthesizes a context-appropriate result $B'$ while preserving unaffected content. It poses two coupled challenges: First, \textit{transformation interpretation} must separate the transferable rule from incidental exemplar properties such as identity, texture, color, and background. Second, \textit{query-conditioned realization} must adapt that rule to the query's structure and context. Because these effects are spatially heterogeneous, transformed objects, interaction regions, secondary effects, and preserved areas may require distinct generation behaviors. Existing methods often encode exemplar relations globally or use image- or layer-level adaptation, capturing only salient effects. As Figure~\ref{fig:task} shows, a model may transfer melting by generating a liquid puddle while leaving the solid volume unchanged---a visually suggestive but incomplete physical transition.

Despite growing interest in physically plausible image editing
\cite{zhao2025risebench,lin2026worldedit,ding2026reedit,
sheng2026ineditbench,xu2026phyeditrealworldobjectmanipulation},
prior work mainly studies text-specified transformations. Whether VICL
models can infer such transformations from visual exemplars, adapt them to
new instances, and generalize to unseen types remains unclear. We introduce
\textit{PhysVICL-74}, a training-and-evaluation benchmark comprising 74
transformation rules, 5,240 source--target image pairs, and nearly 75K
exemplar--query contexts. It supports two complementary protocols.
\textit{Novel-instance transfer} applies seen rules to new objects and
scenes. \textit{Unseen-rule generalization} instead holds out entire rules
during training. Together, they evaluate whether a model follows the
demonstrated rule, adapts it to the query, and preserves rule-irrelevant
content.

Building on these observations, we propose \textit{TransPhy}, a physical-transformation induction and transition-aligned rendering framework that decomposes physically grounded VICL into transformation interpretation and query-conditioned realization. The former captures the transferable rule, while the latter predicts how that rule should manifest in the query, reducing reliance on exemplar-specific appearance. Its understanding pathway predicts textual descriptions of the demonstrated rule and query-specific target state, specifying what transfers and how it manifests while reducing exemplar-specific copying. To realize the predicted transformation, token-wise mixture-of-experts low-rank adaptation (MoE-LoRA) then allows spatial tokens to activate specialized rendering experts. We further develop a training-only \textit{State-Transition Capturer} (STC), which extracts localized transition representations from the vision transformer (ViT) feature differences between $B$ and $B'$. These representations regularize expert routing toward transition-sensitive generation behaviors, encouraging different regions to handle primary changes, secondary effects, and content preservation appropriately. Thus, textual interpretation determines what transformation occurs, while transition-aligned routing controls \textit{how} it is realized across the query image.

Experiments show that TransPhy improves both novel-instance transfer and unseen-rule generalization, producing more complete, physically plausible transformations with competitive perceptual quality and broader visual-relation performance. These results indicate that explicit transformation induction and fine-grained alignment strengthen physically grounded editing and general VICL rule transfer.

Our contributions are summarized as follows:
\begin{itemize}
\item We formulate \textit{physically grounded visual in-context learning}, a setting that requires models to infer a transformation from a visual exemplar and adapt its consequences to the physical properties and context of the query.

\item We introduce \textit{PhysVICL-74}, a dataset and benchmark containing 74 physically grounded transformation rules, 5,240 source--target image pairs, and nearly 75K exemplar--query contexts, with separate evaluation of novel-instance transfer and unseen-rule generalization.

\item We propose \textit{TransPhy}, combining textual rule interpretation with transition-aligned token-wise expert routing to improve physical plausibility and generalization.

\end{itemize}

\section{Related Work}

\paragraph{\textbf{Visual In-Context Learning.}}
In image editing, visual in-context learning (VICL) transfers the transformation demonstrated by an exemplar pair $(A,A')$ to a query image $B$ to synthesize $B'$ \cite{sun2023imagebrushlearningvisualincontext,wang2023incontextlearningunlockeddiffusion}. Existing approaches realize this paradigm through pair-specific optimization \cite{nguyen2023visualinstructioninversionimage,jones2024customizingtexttoimagemodelssingle,lu2025pairedit}, training-free attention or feature manipulation \cite{gu2024analogist,srivastava2024reeditmultimodalexemplarbasedimage,biswas2025pixels}, or learned transfer mechanisms. Representative training-based methods formulate diverse visual tasks as image infilling \cite{li2025visualcloze}, encode relational features with lightweight adapters \cite{gong2025relationadapter,chen2026deltaadapter}, or adapt diffusion transformers through dynamically generated, composed, or routed LoRA modules \cite{song2024lorachange,li2026viral,manor2026lorweb}. Despite substantial progress in visual analogy, existing methods are predominantly developed and evaluated on appearance-, geometry-, or semantics-driven transformations. They do not explicitly address physical rule induction, where successful transfer requires identifying latent state changes and adapting their spatially coupled, scene-dependent consequences to a new query.

\paragraph{\textbf{Physics-Aware Image Editing.}} Recent studies show that physical plausibility remains challenging for multimodal models. PhysBench evaluates reasoning about physical properties, relations, and dynamics \cite{chow2025physbench}, while RISEBench and KRIS-Bench examine broader visual and knowledge-based reasoning \cite{zhao2025risebench,wu2025krisbench}. Physics-aware editing methods and benchmarks further consider latent physical transitions, environmental constraints, causal consistency, physical interactions, and temporal processes \cite{zhao2026staticsdynamicsphysicsawareimage,ding2026reedit,sheng2026ineditbench,han2025unireditbench,lin2026worldedit,pu2025picabench,wu2025chronoedit,guo2026phyeditbench}. However, these settings typically provide the desired effect through a textual instruction or a predefined editing task. The model is asked to execute a known physical transformation, rather than discover it from visual evidence. PhysVICL-74 instead evaluates whether a model can infer the latent physical rule from $(A,A')$ and transfer its scene-dependent consequences to a new query.

\paragraph{\textbf{Unified Multimodal Models.}} Unified multimodal models integrate visual understanding and generation within a shared framework. Janus and Janus-Pro separate understanding and generation encoders \cite{wu2024janus,chen2025januspro}; BAGEL uses shared self-attention with specialized transformer experts \cite{deng2025bagel}; and Show-o2 and JoyAI-Image combine language modeling with generative visual objectives \cite{xie2025showo2,song2026joyai}. Such models provide a basis for physically grounded VICL, as they can jointly compare exemplars, reason about transformations, and generate edited images. Nevertheless, direct generation may still rely on superficial exemplar differences or transfer incidental visual content. This motivates \textit{TransPhy}, which explicitly interprets the demonstrated physical transformation and guides query-specific generation through transition-aligned token-wise expert routing.

\section{Task and Dataset Construction}

\paragraph{Task and Scope.}
Given an exemplar $(A,A')$ and a query image $B$, physically grounded VICL
requires a model to infer the implicit transformation explaining
$A\!\rightarrow\!A'$ and synthesize its query-specific realization $B'$ while
preserving rule-irrelevant content. We use \emph{physically grounded} to
describe visually observable transformations whose qualitative outcomes
depend on material properties, geometry, object interactions, or environmental
conditions. PhysVICL-74 evaluates whether a model transfers the causal
direction and query-appropriate visual consequences of such transformations.
Because multiple outcomes may be physically plausible, each reference target
represents one human-validated plausible realization rather than a unique
physical ground truth. The benchmark therefore does not assess numerical
dynamics or simulator-level physical accuracy.

\paragraph{Transformation Taxonomy.}
We group the 74 rules by their dominant scale and mechanism into three
families: \textit{Scene-Condition Transformations}, comprising Geometric,
Optical, and Temporal Scene Variation; \textit{Mechanically Induced
Transformations}, comprising Action-Induced and Mechanical Response; and
\textit{Material-State Transformations}, comprising Thermophysical and
Reactive Evolution. These seven categories span changes in scene observation, object configuration or deformation, and material state.
For compact reporting, we refer to the Scene-Condition, Mechanically
Induced, and Material-State families as Scene-Level,
Object-Level, and Matter-Level, respectively.

\paragraph{Rule Mining, Construction, and Splits.}
We draw transformation concepts and selected seed images from
RISEBench~\cite{zhao2025risebench}, RE-Edit~\cite{ding2026reedit},
InEdit-Bench~\cite{sheng2026ineditbench},
KRIS-Bench~\cite{wu2025krisbench},
UniREditBench~\cite{han2025unireditbench}, and
WorldEdit~\cite{lin2026worldedit}, rather than directly aggregating their
original image pairs. After merging semantic duplicates and removing ambiguous
or instance-specific edits, we obtain 74 visually observable and transferable
rules. GPT Image completes the corresponding targets and generates additional
rule instances where needed. The resulting benchmark contains 5,240 source--target image pairs
and approximately 75K exemplar--query contexts. Each AABB context
$(A,A',B,B')$ combines two distinct source--target instances governed by the
same rule. We split base image pairs before constructing contexts, ensuring
that no test pair or context containing it appears in training. The protocols
evaluate novel-instance transfer for seen rules and generalization to rules
held out from training.

\paragraph{Target Completion and Verification.}
To mitigate potential generator--reviewer circularity, we do not treat
model-based screening as sufficient. Dedicated human annotators additionally
review rule correctness, physical plausibility, and preservation of
rule-irrelevant content; failed samples are regenerated. Detailed
annotation criteria, review protocols, image provenance, rule mappings, and
representative failure cases are provided in the supplementary material.

\section{Methodology}

\subsection{Overall Framework}
Given an exemplar transformation pair $(A,A')$, a query image $B$, and a fixed, rule-agnostic prompt $p$ (see the supplementary material), 
the model must infer the physical rule demonstrated by $A\!\rightarrow\!A'$ and synthesize its query-specific realization $B'$.
The output should realize the transformation while preserving the identity, spatial layout, and rule-irrelevant content of $B$.

This task involves two levels of reasoning.
At the coarse level, the model must extract a transformation rule that transfers across objects and scenes.
At the fine level, it must adapt the rule to the object properties, spatial structure, and interactions in the query image.
Directly generating $B'$ from the visual exemplar may conflate these two processes, 
causing the model to copy the appearance of $A'$ without transferring the demonstrated rule.

To address this challenge, we propose \textit{TransPhy},
which progressively translates a compact physical rule prior into fine-grained, spatially adaptive rendering decisions.
At the coarse level, the understanding pathway predicts an explicit rule $\hat R$ and a query-specific target-state description $\hat d_{B'}$.
Together, they specify the transformation demonstrated by the exemplar and its expected effect on the query.
At the fine level, the generation pathway employs token-wise MoE-LoRA, allowing different spatial tokens to invoke different low-rank rendering experts.
We further introduce a training-only \textit{State-Transition Capturer} (STC), which derives token-level transition targets from ViT feature differences between $(B,B')$ and uses them to supervise expert routing.

TransPhy follows a coarse-to-fine path from
\textit{physical-rule induction} to
\textit{transition-aligned expert rendering}.
We instantiate it on BAGEL.

\begin{figure*}[t]
    \centering
    \includegraphics[width=\textwidth,trim=122.25bp 294.75bp 224bp 114.5bp,clip]{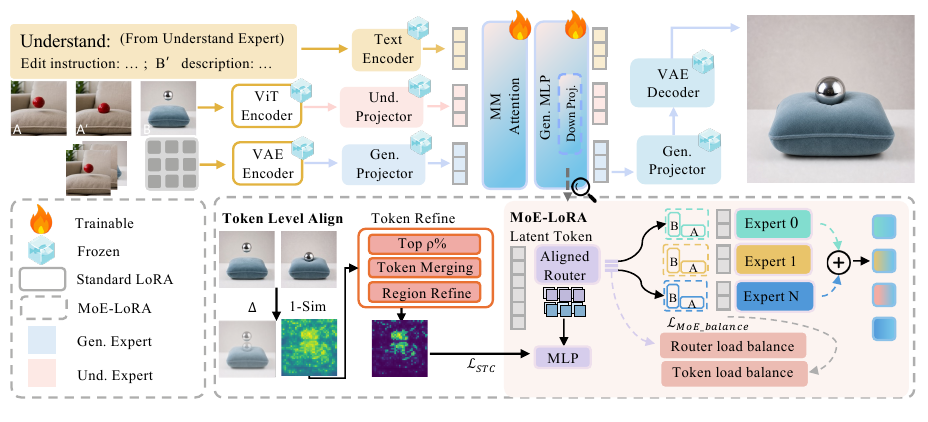}
    \caption{Overview of TransPhy on BAGEL. The understanding and generation pathways encode the exemplar--query context; ViT-derived transition evidence aligns token-level routing, while MoE-LoRA experts render the inferred physical rule.}
    \label{fig:method}
    \vspace{-13pt}
\end{figure*}

\subsection{Progressive Physical Rule Induction and Rendering}

\paragraph{Coarse-Grained Physical Rule Prior.}
BAGEL employs a shared multimodal transformer that supports autoregressive understanding and DiT-based visual generation, 
providing a unified interface between explicit rule inference and image synthesis.
Given the interleaved context $(A,A',B,p)$, the understanding pathway first predicts the demonstrated rule $\hat R$ and a query-specific target-state description $\hat d_{B'}$.
We then combine these textual intermediates with the visual representations to construct the generation context:
\begin{equation}
c=[c_A,c_{A'},c_B,T_p(\hat R,\hat d_{B'})],\qquad
\hat v=v_\phi(z_t,t,c),
\end{equation}
where $c_I$ denotes BAGEL's multimodal token representation of image $I$, $T_p$ serializes the task prompt and predicted intermediates into text tokens, $z_t$ is the noisy visual latent at timestep $t$, and $v_\phi$ predicts the visual training target.

Together, $\hat R$ and $\hat d_{B'}$ form a compact, coarse-grained physical rule prior.
This representation explicitly summarizes the transformation to be transferred and describes its expected realization on the query.
Its primary role is to provide coarse-grained semantic constraints.
Conditioning generation on this explicit interpretation reduces the shortcut of copying the appearance of $A'$ and provides a stable semantic condition for subsequent fine-grained rendering.

\paragraph{Fine-Grained Token-Wise Expert Rendering.}
The physical rule prior describes the global semantics of the target state, but state-transition editing typically exhibits substantial spatial heterogeneity.
The transformed object, interaction areas, secondary effects, and preserved content may require different rendering behaviors.
We freeze the BAGEL backbone and apply standard LoRA
\cite{hu2021lora} to the understanding pathway and most generation projections.
However, standard LoRA applies the same low-rank update to every spatial token and therefore cannot adapt its computation according to the spatial role of each region.

Motivated by mixtures of LoRA experts \cite{wu2024mole}, we replace the
down-projection of the generation MLP with token-wise MoE-LoRA:
\begin{equation}
y_i
=
Wx_i+
\frac{\alpha}{r}
\sum_{e=1}^{E}
g_{i,e}U_eV_ex_i,
\end{equation}
where $x_i$ is the $i$-th generation token, $W$ is the frozen pretrained projection, $U_eV_e$ is the rank-$r$ update of expert $e$, $\alpha$ controls the adapter scale, and $E$ is the number of experts.
The router independently computes the expert weights for each token using
sparse top-$k$ routing:
\begin{equation}
g_i=
\operatorname{TopKSoftmax}
\left(W_{\rm r}x_i/\tau,k\right),
\end{equation}
where $W_{\rm r}$ is the router projection, $\tau$ is the routing temperature, and $k$ is the number of activated experts.

Because $g_i$ is predicted separately for every token, different spatial positions can invoke different low-rank updates.
The coarse-grained physical rule prior can therefore be progressively realized through fine-grained, spatially adaptive expert rendering.
Nevertheless, token-wise MoE-LoRA provides only the capacity for fine-grained rendering; 
it does not guarantee that the router will organize experts according to the spatial effects of the transformation.
When trained only with the image-generation objective, the router may instead specialize according to generic appearance cues such as color, texture, or object category.
We therefore introduce fine-grained transition alignment.

\subsection{Fine-Grained Transition Alignment}

During training, MoE-LoRA experts learn their rendering behavior under the guidance of the token-wise router.

To provide token-level supervision, STC aligns the router's spatial perception with localized 
transition evidence extracted by a frozen ViT.

\paragraph{ViT-Derived Transition Targets.}
We extract semantic features from $(B,B')$ using BAGEL's frozen ViT
encoder, initialized from \textsc{SigLIP2-so400m/14}~\cite{deng2025bagel}.
Their differences localize the visible effects of the transformation.
Let $u_j$ and $u'_j$ denote the frozen ViT embeddings of $(B,B')$ at spatial position $j$.
We define the initial fine-grained local transition response as
\begin{equation}
s_j
=
1-
\frac{u_j^\top u'_j}
{\|u_j\|_2\|u'_j\|_2}.
\end{equation}
A larger $s_j$ indicates a stronger semantic change between the source and target states.

We retain the top $\rho\%$ transition-responsive tokens and refine these sparse candidates through connected-component merging and feature-similarity refinement~\cite{shang2026llavaprumergeadaptivetokenreduction,liu2024parpromptawaretokenreduction}.
Details of token selection and refinement are provided in the supplementary material.
The resulting transition target $q$ is resized onto the VAE token grid of $B'$ using nearest-neighbor interpolation.
Compared with raw pixel differences, ViT feature differences are less sensitive to color shifts, generation noise, and local texture variations, providing a more stable token-level supervision signal.

\paragraph{Router Alignment.}
Let $a_i\in\mathbb R^E$ denote the pre-softmax router logits for the
$i$-th generation token of $B'$. A two-layer prediction head
$f_{\rm STC}:\mathbb R^E\rightarrow\mathbb R$ converts the
$E$ expert-routing logits into a scalar transition score. We align
its sigmoid response with the resized ViT-derived target:
\begin{equation}
\mathcal L_{\rm STC}
=
\frac{1}{|\Omega_{B'}|}
\sum_{i\in\Omega_{B'}}
\left(
\sigma(f_{\rm STC}(a_i))-q_i
\right)^2,
\end{equation}
where $\Omega_{B'}$ indexes the VAE token positions of $B'$.

This objective encourages the internal routing representation to distinguish transition-responsive positions from relatively stable regions, without assigning a predefined semantic identity to any expert.
The rendering behavior of each expert remains learned through the image-generation objective.
STC therefore aligns where expert responses should vary, rather than prescribing which expert must represent a particular transformation.

To reduce expert collapse, we additionally introduce a standard MoE
load-balancing objective:
\begin{equation}
\mathcal L_{\rm bal}
=
E\sum_{e=1}^{E}\bar p_e\bar\ell_e,
\end{equation}
where $\bar p_e$ is the mean routing probability of expert $e$ and $\bar\ell_e$ is the fraction of tokens assigned to it.
The generation objective learns the rendering behavior of the experts, $\mathcal L_{\rm STC}$ makes the router sensitive to the spatial effects of the transformation, and $\mathcal L_{\rm bal}$ encourages non-collapsed expert specialization.

\subsection{Staged Training Objective}

We train TransPhy in two stages: coarse-grained physical-rule understanding
and fine-grained rendering. In Stage~1, we freeze the BAGEL backbone and
optimize a rank-16 LoRA with autoregressive supervision
\begin{equation}
y=[R,d_{B'}],
\end{equation}
where $R$ is the annotated transformation rule and $d_{B'}$ is the
query-specific target-state description. This stage constrains the
understanding pathway to a stable ``rule--target description'' output format,
while learning to extract the demonstrated rule and adapt its expected effect to the query.

In Stage~2, we freeze Stage~1 and condition generation on its predicted textual intermediates.
We employ rank-32 generation adapters and four MoE-LoRA experts, and jointly optimize visual generation, 
transition alignment, and expert load balancing:
\begin{equation}
\mathcal L_{\rm render}
=
\mathbb E_t
\left[
\|v-v_\phi(z_t,t,c)\|_2^2
\right]
+
\lambda_{\rm STC}\mathcal L_{\rm STC}
+
\lambda_{\rm bal}\mathcal L_{\rm bal},
\end{equation}
where $v$ is the training target, and $\lambda_{\rm STC}$ and $\lambda_{\rm bal}$ control the transition-alignment and load-balancing terms, respectively.

This staged optimization establishes a coarse-to-fine learning process.
Stage~1 constructs a stable physical rule prior, while Stage~2 translates this prior into spatially adaptive token-wise expert rendering.
By assigning different rendering experts according to the query content and transition evidence, 
TransPhy improves the spatial precision and controllability of physically grounded state-transition editing.

\section{Experiments}

\begin{figure*}[t]
\centering
\includegraphics[
    width=\textwidth,
    page=1,
    trim=42bp 159bp 42bp 116bp,
    clip
]{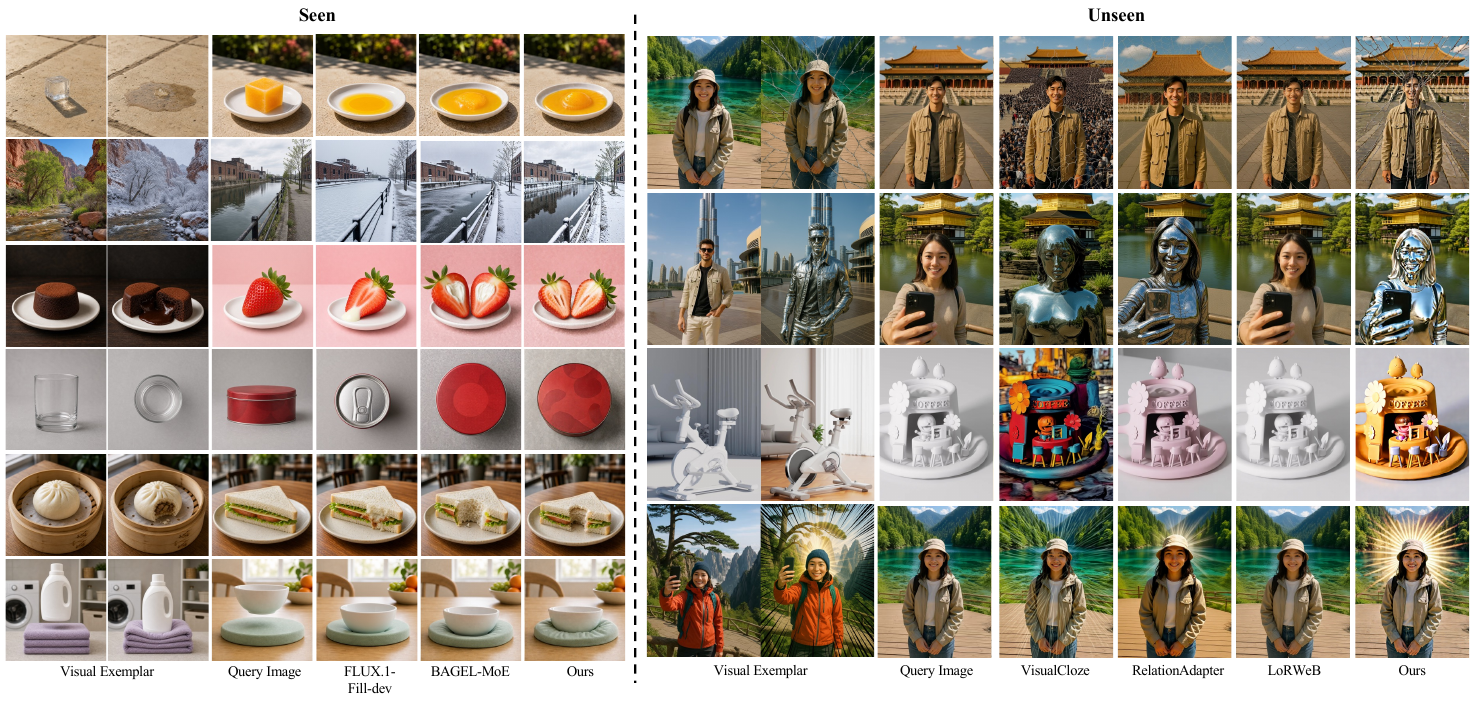}
\caption{\textbf{Qualitative comparison on seen and unseen transformations.}
Left: new instances of seen transformations. Right: unseen transformations.}
\label{fig:qualitative-results}
\end{figure*}

\begin{table*}[t]
\centering
\scriptsize
\setlength{\tabcolsep}{4.0pt}
\caption{\textbf{Seen-rule transfer to novel instances across three rule
families.} Scene-Level, Object-Level, and Matter-Level denote Scene-Condition,
Mechanically Induced, and Material-State transformations, respectively.
TA, CP, and RP are scored by \mbox{GPT-5.6}; CLIP-D and LPIPS are objective
metrics. Results are macro-averaged over rules within each family.}
\label{tab:seen}
\begin{tabular*}{\textwidth}{@{\extracolsep{\fill}}lccccccccc@{}}
\hline
& \multicolumn{3}{c}{\textbf{FLUX.1-Fill-dev}} & \multicolumn{3}{c}{\textbf{BAGEL-MoE}} & \multicolumn{3}{c}{\textbf{TransPhy}} \\
\cline{2-4}\cline{5-7}\cline{8-10}
Metric
& Scene-Level
& Object-Level
& Matter-Level
& Scene-Level
& Object-Level
& Matter-Level
& Scene-Level
& Object-Level
& Matter-Level \\
\hline
TA$\uparrow$ & 3.45 & 3.12 & 3.33 & 3.22 & 3.03 & 3.63 & \textbf{3.57} & \textbf{3.39} & \textbf{3.73} \\
CP$\uparrow$ & 3.23 & 3.18 & 3.41 & 3.12 & 3.57 & 3.23 & \textbf{3.43} & \textbf{3.78} & \textbf{3.67} \\
RP$\uparrow$ & 3.20 & 3.67 & 3.09 & 3.33 & 3.21 & 3.28 & \textbf{3.47} & 3.36 & 3.25 \\
CLIP-D$\uparrow$ & 0.20 & 0.25 & 0.25 & 0.21 & 0.27 & 0.29 & \textbf{0.23} & 0.26 & 0.27 \\
LPIPS$\downarrow$ & 0.39 & 0.36 & 0.26 & 0.40 & 0.36 & 0.26 & \textbf{0.31} & \textbf{0.31} & \textbf{0.24} \\
\hline
\end{tabular*}
\vspace{-8pt}
\end{table*}

\begin{table}[t]
\centering
\scriptsize
\setlength{\tabcolsep}{1.7pt}
\caption{\textbf{Generalization to transformations unseen during TransPhy
training.} Results are reported as PhysVICL-74/Relation252K over 17
held-out PhysVICL-74 rules and 20 sampled Relation252K rules, respectively.}
\label{tab:heldout}
\resizebox{\columnwidth}{!}{
\begin{tabular}{lccccc}
\hline
Method
& TA$\uparrow$
& CP$\uparrow$
& RP$\uparrow$
& CLIP-D$\uparrow$
& LPIPS$\downarrow$ \\
\hline
FLUX.1-Fill-dev
& 2.63/1.60 & 3.45/3.71 & 3.22/1.79 & 0.24/0.26 & 0.51/0.56 \\

BAGEL-MoE
& 2.55/2.16 & 3.60/2.96 & 3.24/1.90 & 0.24/0.23 & 0.54/0.61 \\

\rowcolor[gray]{0.94}
TransPhy
& \textbf{2.91}/2.34
& \textbf{3.75}/3.05
& 3.28/\textbf{2.90}
& \textbf{0.31}/0.28
& \textbf{0.48}/\textbf{0.50} \\
\hline

RelationAdapter
& 2.60/3.21 & 3.17/3.88 & 3.34/2.82 & 0.20/0.37 & 0.52/0.52 \\

VisualCloze
& 2.76/2.10 & 3.15/3.41 & 3.04/2.45 & 0.25/0.22 & 0.51/0.54 \\

LoRWeB
& 1.41/2.16 & 3.71/2.63 & 3.44/2.43 & 0.12/0.26 & 0.58/0.53 \\
\hline
\end{tabular}
}
\end{table}

\subsection{Settings}
We instantiate TransPhy on BAGEL-7B-MoT and freeze the backbone.
The understanding and generation stages use LoRA ranks 16 and 32, respectively; the generation adapter contains four MoE-LoRA experts with top-1 routing, and STC retains the top 15\% transition-responsive ViT tokens.
We optimize both stages with AdamW and bfloat16 precision at a learning rate of $1\times10^{-4}$ with 100 warmup steps.
The understanding stage is trained for 8{,}000 steps and the generation stage for 20{,}000 steps.
Training uses full-shard FSDP on four A100 GPUs, a per-GPU batch size of 1, and two-step gradient accumulation, yielding an effective batch size of 8.
Further implementation details are provided in the supplementary material.

\subsection{Benchmark}
PhysVICL-74 contains 74 transition rules, 5,240 source--target image pairs, and approximately 75K AABB contexts.
The evaluation benchmark combines three disjoint sources: novel instances of 57 PhysVICL-74 rules represented in training, 17 PhysVICL-74 rules entirely unseen during training, and 20 sampled Relation252K rules not included in our training, with four sampled pairs per rule in each source.
For the first split, both the exemplar pairs and queries are image-disjoint from training; for the latter two, the rules and all associated images are held out.
We split base image pairs before constructing AABB contexts, ensuring that no test pair or context containing it appears in training.

\subsection{Baseline Methods}
For novel instances of seen rules, we compare TransPhy with \textit{FLUX.1-Fill-dev} and \textit{BAGEL-MoE}.
FLUX.1-Fill-dev is trained on the same PhysVICL-74 split, whereas BAGEL-MoE shares our backbone, adapters, training data, and inference settings but removes STC-based router alignment.
For unseen rules, we additionally compare with RelationAdapter~\cite{gong2025relationadapter}, VisualCloze~\cite{li2025visualcloze}, and LoRWeB~\cite{manor2026lorweb}, using their released checkpoints, official inference settings, and native input formats.
All methods receive the same exemplar--query semantics.
Because RelationAdapter and LoRWeB do not publicly disclose detailed training-data splits, exact training-data alignment cannot be verified.

\subsection{Evaluation Metrics}
We evaluate with five complementary metrics. 
We adopt CLIP Directional Similarity (CLIP-D)~\cite{manor2026lorweb} using \textsc{OpenAI CLIP
ViT-L/14} and Learned Perceptual Image Patch Similarity (LPIPS)~\cite{zhang2018unreasonable}. 
Following VLM-based protocols for visual-analogy and knowledge-aware editing~\cite{manor2026lorweb,lin2026worldedit}, 
GPT-5.6 scores \textit{Transition Accuracy} (TA), \textit{Content Preservation} (CP), and \textit{Rule Plausibility} (RP) on a 0--4 scale, with higher scores being better. 
These three complementary metrics assess transition fidelity, query preservation, and physical consistency, respectively. Full definitions, prompts, and aggregation details are provided in the supplementary material.
We additionally conduct an anonymized, criterion-specific two-alternative forced-choice (2AFC) user study; details are provided in the supplementary material.

\vspace{-2pt}
\paragraph{Quantitative Evaluation.}
As shown in Table~\ref{tab:seen}, TransPhy achieves clear overall
gains over BAGEL-MoE on seen transformations, increasing Object-Level TA
from 3.03 to 3.39 and Matter-Level CP from 3.23 to 3.67. On unseen
transformations (Table~\ref{tab:heldout}), it outperforms the matched
BAGEL-MoE baseline on every metric in both settings and ranks first in six
of ten metric--setting combinations among all compared methods. These results
demonstrate strong and balanced generalization across transition fidelity,
query preservation, physical plausibility, and perceptual fidelity.

\paragraph{Qualitative Evaluation.}
Figure~\ref{fig:qualitative-results} compares seen and unseen
transformations.
On seen rules (left), TransPhy faithfully realizes
diverse physical changes while preserving query geometry;
FLUX.1-Fill-dev captures coarse target appearances, whereas
BAGEL-MoE spreads or under-applies edits.
On unseen rules (right), our method transfers material,
representation, and lighting effects with limited background
changes.
RelationAdapter often under-applies the effect,
VisualCloze alters background structures, and LoRWeB leaves
queries nearly unchanged, showing that TransPhy better
balances complete transfer and query preservation.

\section{Ablation Studies}
To investigate the impact of different components of our framework, we conduct the following ablation studies.

\noindent\textbf{(1) Staged Rule Understanding and Expert Alignment.}
We ablate the two central training components in TransPhy.
\textit{W/o Stage~1} skips dedicated rule-understanding training and
conditions the generation pathway on a single fixed instruction shared
across instances.
\textit{W/o STC alignment} retains Stage~1 but removes
$\lambda_{\rm STC}\mathcal L_{\rm STC}$ from generation training while
keeping the generation and load-balancing objectives unchanged.
This directly tests whether explicit rule internalization and STC-guided
expert routing are necessary for generalizable transformation transfer.

\noindent\textbf{(2) Number of Rendering Experts.}
We vary the number of MoE-LoRA rendering experts as $E\in\{4,8,16\}$ and use $E=4$ in the main experiments.
This ablation examines whether a larger expert pool provides additional capacity for modeling diverse visual transformations.

\begin{table}[ht]
\centering
\scriptsize
\setlength{\tabcolsep}{2.2pt}
\caption{\textbf{Ablation of training design and expert count.} The full model uses staged training, STC alignment, and four experts. The first two variants modify the training design; the last two change only the expert count.}
\label{tab:training-expert-ablation}
\begin{tabular}{lccccc}
\hline
Variant & TA$\uparrow$ & CP$\uparrow$ & RP$\uparrow$ & CLIP-D$\uparrow$ & LPIPS$\downarrow$ \\
\hline
Full model ($E=4$) & 3.25 & 3.46 & 3.36 & 0.27 & 0.31 \\
W/o Stage~1 & 1.78 & 3.08 & 1.87 & 0.21 & 0.39 \\
W/o STC align. & 3.18 & 3.24 & 2.95 & 0.26 & 0.37 \\
$E=8$ & 3.28 & 3.46 & 3.66 & 0.29 & 0.29 \\
$E=16$ & 3.34 & 3.52 & 3.74 & 0.29 & 0.30 \\
\hline
\end{tabular}
\end{table}

\noindent\textbf{(3) Transition-Token Selection Sensitivity.}
To evaluate STC token localization, GPT-5.6 annotates the transformation-affected ViT-grid tokens $M$ for a fixed subset of five source--target pairs per transformation rule.
The masks are audited by Qwen3-VL-32B, with stratified manual spot checks and corrections; criteria are provided in the supplement.
We retain the top $\rho\in\{5,15,30\}\%$ candidates by ViT cosine difference and apply the same merging and refinement steps to obtain $S_\rho$.
We report token-level precision, recall, and F1 between $S_\rho$ and $M$, averaged across samples.

\begin{table}[ht]
\centering
\small
\setlength{\tabcolsep}{4pt}
\caption{\textbf{Transition-token selection quality.} Each $\rho$ is evaluated on the fixed subset after connected-component merging and feature-similarity refinement.}
\label{tab:stc-token-ablation}
\begin{tabular}{lccc}
\hline
Metric & $\rho=5\%$ & $\rho=15\%$ & $\rho=30\%$ \\
\hline
Token precision$\uparrow$ & 84.18 & 69.19 & 41.65 \\
Token recall$\uparrow$ & 28.25 & 65.03 & 72.57 \\
Token F1$\uparrow$ & 42.30 & 67.05 & 52.92 \\
\hline
\end{tabular}
\end{table}

\begin{figure}[ht]
\centering
\includegraphics[
    width=\columnwidth,
    trim=48bp 453bp 520bp 111bp,
    clip
]{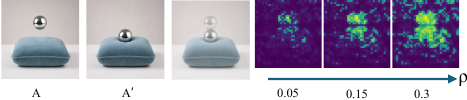}
\caption{\textbf{Transition-token visualization across $\rho$.}}
\label{fig:rho-visualization}
\end{figure}

\begin{figure}[!t]
\centering
\includegraphics[width=\columnwidth,trim=76bp 400bp 462bp 115bp,clip]{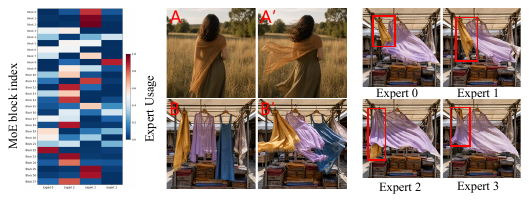}
\caption{\textbf{Expert routing visualization.}
Left: layer-wise expert-routing heat map.
Right: results obtained by hard-routing all tokens through
experts~0--3.
}
\label{fig:expert-routing-case}
\vspace{-12pt}
\end{figure}

\noindent\textbf{(4) Expert Intervention and Interpretability.}
Holding all other settings fixed, we route all generation tokens through a
single expert $e\in\{0,1,2,3\}$ and compare the outputs with natural
layer-wise routing in Figure~\ref{fig:expert-routing-case}.

\noindent\textbf{Results \& Analysis.}
\textbf{(1)} Table~\ref{tab:training-expert-ablation} shows that removing
Stage~1 causes the largest drops in TA (3.25 to 1.78) and RP (3.36 to 1.87),
confirming that rule internalization is central to transfer.
Removing STC alignment reduces RP to 2.95 and increases LPIPS from 0.31 to
0.37, indicating that expert alignment improves plausible and
spatially faithful rendering.
\textbf{(2)} Increasing $E$ from 4 to 16 yields modest gains: $E=8$ gives
the lowest LPIPS, whereas $E=16$ performs best on TA, CP, and RP.
This suggests the bottleneck may lie in the base model rather than
expert capacity.
\textbf{(3)} Table~\ref{tab:stc-token-ablation} and
Figure~\ref{fig:rho-visualization} show that $\rho=5\%$ favors precision but
has low transition-region recall, whereas $\rho=30\%$ improves recall at the
cost of noisy selections. $\rho=15\%$ achieves the highest F1 of 67.05,
validating a moderate selection ratio.
\textbf{(4)} Figure~\ref{fig:expert-routing-case} shows that routing varies
across layers and that forced experts produce distinct localized effects.
This provides evidence of emergent, non-redundant expert
specialization rather than fixed semantic roles.
Taken together, these complementary results provide consistent evidence
for the effectiveness of staged induction and transition-aligned expert
routing.

\section{Conclusion}
We formulated physically grounded VICL, where models infer a transformation
rule from an exemplar and apply it to a new query.
We introduced PhysVICL-74, covering 74 rules and
approximately 75K exemplar--query contexts, with protocols for novel-instance
transfer and unseen-rule generalization.
We proposed TransPhy, combining 
physical-rule induction with transition-aligned token-wise expert adaptation
for fine-grained and physically plausible synthesis.
Extensive experiments show strong performance on physically grounded
transformations and improved generalization to unseen rules over existing
VICL methods.
We hope these contributions support future study of visual rule induction
and context-aware image editing.

\bibliography{transphy}

\end{document}